\documentclass{article}
\PassOptionsToPackage{numbers,sort&compress}{natbib}
\usepackage[preprint]{neurips_2026}
\usepackage[T1]{fontenc}
\usepackage[utf8]{inputenc}
\usepackage{times}
\usepackage{amsmath,amssymb,booktabs,array,tabularx,tikz,listings}
\usepackage{xcolor}
\usepackage{microtype,url,hyperref,graphicx}
\usepackage[labelsep=period]{caption}
\usetikzlibrary{arrows.meta,shapes.geometric}
\hypersetup{hidelinks}
\newcommand{\gain}{\ensuremath{\times}}

\title{Towards Certificate-Driven Software Porting:\\
  A Self-Improving Agentic Harness\\
  for Scientific Program Optimization}
\workshoptitle{Machine Learning and the Physical Sciences}

\author{
  Piyush Jha\textsuperscript{1,2}
  \qquad
  Aishik Ghosh\textsuperscript{2,3}
  \qquad
  Vijay Ganesh\textsuperscript{1}
  \\[1ex]
  \textsuperscript{1}School of Computer Science, Georgia Institute of Technology, USA \\
  \textsuperscript{2}School of Physics, Georgia Institute of Technology, USA \\
  \textsuperscript{3}Lawrence Berkeley National Laboratory, USA \\
  \texttt{piyush.jha@gatech.edu,}
  \texttt{aishikghosh@physics.gatech.edu,} \\
  \texttt{vganesh@gatech.edu}
}

\begin{document}
\maketitle

\begin{abstract}
The upgrade and rewriting of large scientific codebases has traditionally been a major challenge. While evolutionary search with large language models (LLMs) can port and accelerate legacy code, repair feedback in prompts alone does not prevent subsequent candidates from repeating the same errors. We introduce Certificate-Driven Evolutionary Search (CDES), which extends evolutionary search with enforceable restrictions derived from failed candidates, recorded as \textit{certificates} of assumptions, checker evidence, and justified restrictions. Its control logic enforces these restrictions through rejection, backtracking, and targeted repair while preserving compatible edits. 
We apply CDES to CPU-to-GPU translation of two particle-simulation functions from the Geant4 toolkit, evaluated with a harness that goes beyond unit tests to combine formal checks, numerical comparisons, physics checks, and GPU safety tests. Generated implementations achieve 13.78\gain\ and 23.54\gain\ function-level speedups over CPU code, including data conversion and transfers; for one function, GPU throughput exceeds an expert implementation by 14.9\%, reaching 16.1\% when complementary components are combined. In an ablation over execution settings, certificate feedback increases the fraction of candidates passing required correctness checks from 55\% to 90\%.
\end{abstract}

\section{Introduction}

Scientific software is often developed and extended by the community over several decades, making it challenging to rewrite the entire codebase whenever new hardware acceleration paradigms emerge. Simulators are of particular interest because they consume a substantial amount of computational time, for instance in physics and astrophysics research~\cite{ATLAS:2021pzo,ATLAS:2022jhk,Xu_2016,Liu:2023lnn}. There is therefore great interest in faster implementations of simulation software that preserve computational and physical specifications. Geant4 is a particle simulation toolkit widely used for detector simulation in particle physics experiments and for applications in medical imaging~\citep{geant4,geant42006,geant42016}. While porting the entirety of Geant4 to GPU-accelerated code is a herculean task, the Celeritas project has taken a significant step by designing expert-written GPU implementations for particle transport~\citep{celeritas}. In this proof-of-concept study, we explore how automated optimization can complement these efforts by improving existing implementations and accelerating functions for which no expert-written GPU implementations exist. For this approach to scale to the entire codebase, we assume that selected functions can be optimized independently, and that speeding up individual components will reduce overall simulation time. We evaluate function-level performance here, leaving full-simulation gains to future work. 

CDES builds on state-of-the-art evolutionary program search, where systems such as FunSearch, AlphaEvolve, ERA, and MadEvolve repeatedly generate, evaluate, and select programs~\citep{funsearch,alphaevolve,era,madevolve}. Our focus is on how information from failed candidates can be converted into enforceable restrictions on subsequent search. The certificate database provides feedback to the mutator LLM, but CDES does not rely on prompt compliance alone. Its control logic separately enforces validated restrictions on candidate choices, rejecting conflicting choices and guiding the repair process. This design works with frontier LLMs through API calls and requires neither access to model weights nor additional training. It draws on symbolic feedback in RLSF~\citep{rlsf}, constrained reinforcement learning in CDRL~\citep{cdrl}, and conflict learning for program construction in Neo~\citep{neo}.

We make two contributions. First, a \textbf{certificate-driven evolutionary search mechanism} converts conflicts into verified restrictions and enforces them through targeted rejection and repair while preserving compatible changes. Second, a \textbf{scientific optimization and evaluation harness} integrates this mechanism with tool selection and correctness checks to translate CPU programs into CUDA, NVIDIA's GPU programming platform. The search is self-improving because accumulated certificates change which future modifications are permitted, without retraining the LLM.

\section{Certificate-Driven Evolutionary Search}

\begin{figure}[t]
  \centering
  \includegraphics[width=0.85\linewidth]{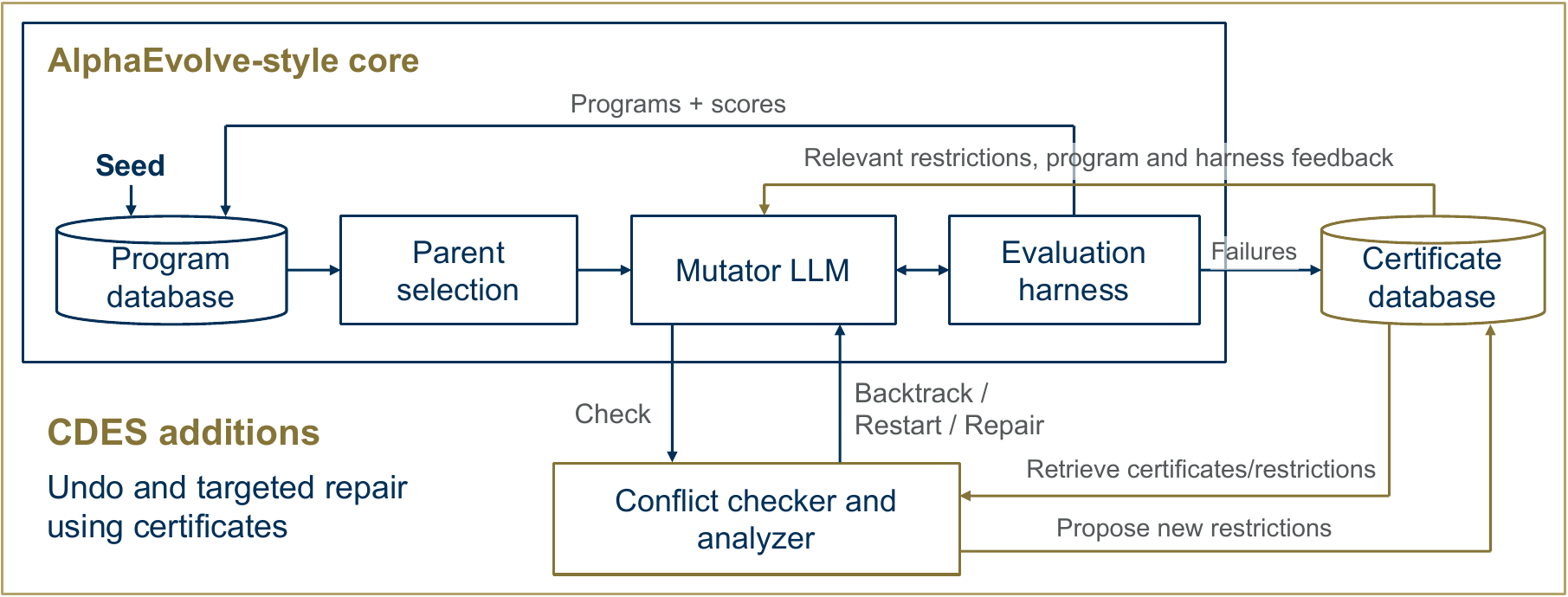}
  \caption{CDES extends evolutionary search with certificate-driven control. The certificate database supplies restrictions and feedback to the mutator LLM and evidence to the conflict checker and analyzer. CDES enforces validated restrictions during generation and repair. Gold highlights CDES additions.}
  \label{fig:architecture}
\end{figure}

CDES extends an AlphaEvolve-style core implemented with ShinkaEvolve~\citep{shinka} (Figure~\ref{fig:architecture}). A seed initializes the program database. Parent selection chooses a stored program, and the mutator LLM proposes code or execution settings. The evaluation harness returns programs and scores to the program database and failures to the certificate database. CDES adds backtracking, restarting, and targeted repair.

\textbf{Detection and certification.} The evaluation harness detects requirement violations. The conflict checker and analyzer reviews unfinished candidates using stored certificates and symbolic checks of dependencies between code components. A certificate records the conflicting choices, the violated requirement, and supporting assumptions and evidence. Proposed restrictions are checked before being used to block the same conflict in later candidates.

\textbf{Undo and targeted repair.} CDES control logic rejects conflicting combinations, adds required components, and repairs choices while preserving compatible edits. Conflict-driven clause learning (CDCL) SAT solvers determine whether Boolean choices satisfy logical constraints. They learn restrictions from conflicts and backjump to earlier decisions~\citep{cdcl}. CDES similarly backtracks, restarts, or repairs candidates, without claiming that no solution exists. Unlike prompt feedback, validated restrictions are enforced even when the LLM repeats a conflicting proposal.


Related ideas appear in program analysis and synthesis. Prior work uses program summaries and simplified models to guide search~\citep{atlas,syngar}, identifies conditions under which rewrites preserve behavior~\citep{aliveinfer}, and aligns states across program executions~\citep{alignment}. Other approaches check unfinished programs~\citep{synquid} or infer properties preserved across loop iterations~\citep{hornice}. CDES draws on these directions to guide tool requests and the construction of reusable restrictions, rather than implementing each method directly.

\section{Evaluation Harness}
\label{sec:correctness}

Geant4 source and specifications define expected behavior, with Celeritas as an expert performance reference where available. An evaluator LLM selects and orders tools, while a fixed policy mandates checks described below~\citep{llmmodulo}. Each evaluation result records a hash identifying the exact candidate source code tested. Failed or unresolved mandatory checks block acceptance and enter the certificate database. Separate validation suites test additional workload sizes, fresh random-number sequences, and inputs near numerical decision boundaries. Their results are withheld from the corresponding search runs rather than used to guide proposals.

\textbf{Formal analysis.} The Lean package~\citep{lean} verifies selected requirements, called contracts, rather than whole-program equivalence. GPU assignment proofs ensure each input is processed exactly once and retains its assigned random-sequence identifier. Restricted code generation translates the GPU work-assignment rules checked in Lean into executable code. CBMC checks C array indexing within bounded executions~\citep{cbmc}. Physlib/HepLean and Mathlib provide reusable physics and mathematics proofs~\citep{physlib,heplean,mathlib}. Alive2 and VOLTA check code-fragment equivalence in separate demonstrations~\citep{alive2,volta}. These guarantees cover selected requirements and fragments, not complete GPU programs.

\textbf{Execution-based validation.} Differential tests compare reference and candidate outputs on matching inputs within fixed numerical tolerances. Where contracts require, they also check exact output counts and random-number consumption. Physics checks test conservation laws and valid output ranges. Compute Sanitizer detects GPU memory-access and synchronization errors~\citep{sanitizer}. Statistical tests compare selected output distributions from independent samples. Nsight Compute profiles GPU execution~\citep{nsight}. Acceptance means passing required checks within their stated domains, not universal program equivalence.

\section{Experimental Design}

We study two complementary Geant4 functions as case studies, translating CPU-only C++ functions rather than the complete simulation. The Compton function computes particle energies and directions after a photon scatters from an electron. The Fermi function assigns energy and motion to fragments of an atomic nucleus. Optimization must satisfy the evaluation criteria in Section~\ref{sec:correctness}.

We chose Compton because Celeritas has an expert GPU implementation, and Fermi because, to our knowledge, it has no matching GPU implementation already available. This lets us test both improving on existing GPU code and generating GPU code where none is provided by Celeritas. Celeritas therefore serves as our expert reference~\citep{celeritas} for the first function. The Fermi case also reduces reliance on a comparison where the LLM may have seen the Celeritas code during pretraining.

We benchmark the CPU reference, CDES-generated GPU code, and Celeritas where available, using approximately 250,000 inputs on a 10 GB partition of an NVIDIA H100 GPU. Comparisons within each table use matching inputs and test programs with repeated measurements. Table~\ref{tab:timing} sums separately measured data conversion, transfers, and GPU computation. CPU timings include allocation, whereas GPU timings use preallocated buffers. Table~\ref{tab:hybrid} directly measures transfers and computation in a shared benchmark with different supporting code and launch settings. Its GPU buffers are also preallocated. Absolute times should therefore be compared within each table. Speedup is reference time divided by generated-code time, and throughput is inputs processed per second. Both tables exclude search and evaluation runtime. We use GPT-5.6-Sol with medium reasoning, with total model-call costs under \$10.

\begin{table}[t]
  \caption{Generated CUDA achieves \textbf{13.78\gain} and \textbf{23.54\gain} speedups over standalone CPU functions under the stated timing protocol. Times are in milliseconds. GPU times sum separately measured conversion, transfers, and computation, not full-simulation runtime.}
  \centering\small
  \setlength{\tabcolsep}{4pt}
  \renewcommand{\arraystretch}{0.95}
  \begin{tabular}{lrrr}
    \toprule
    Function & Geant4 CPU & CDES-generated CUDA & Speedup \\
    \midrule
    Compton & 30.162 & \textbf{2.189} & \textbf{13.78\gain} \\
    Fermi & 340.090 & \textbf{14.447} & \textbf{23.54\gain} \\
    \bottomrule
  \end{tabular}
  \label{tab:timing}
\end{table}

\section{Results}

\textbf{Performance.} CDES-generated GPU code achieves \textbf{13.78\gain} and \textbf{23.54\gain} speedups over the CPU versions under the standalone timing protocol (Table~\ref{tab:timing}). In the shared Compton benchmark, GPU throughput is \textbf{14.9\%} higher than Celeritas, while execution time including transfers is approximately \textbf{3.0\%} lower (Table~\ref{tab:hybrid}).

\textbf{Optimization patterns.} Generated code reuses earlier calculations and simplifies repeated work. In Compton, it keeps a value that Celeritas computes inside a helper (Figure~\ref{fig:code}). In Fermi, it replaces selected power calculations with multiplication, retaining the original calculation where rounding could change a decision. Reported speedups measure complete functions, not individual changes.

\begin{figure}[!ht]
  \centering
  \begin{minipage}[t]{.49\linewidth}
    \textbf{Celeritas helper}\par
\begin{lstlisting}
// Compute from the cosine
costheta = 1 - t;
sintheta = sqrt(1 - costheta*costheta);
\end{lstlisting}
  \end{minipage}\hfill
  \begin{minipage}[t]{.49\linewidth}
    \textbf{CDES-generated code}\par
\begin{lstlisting}
// Reuse an earlier calculation
if (sint2 < 0) sint2 = 0;
sinTeta = sqrt(sint2);
\end{lstlisting}
  \end{minipage}
  \caption{Avoiding repeated work in Compton. Celeritas computes the squared sine inside its helper. CDES-generated code reuses this quantity from an earlier calculation. The conditional prevents a negative input to the square root. Snippets are shortened.}
  \label{fig:code}
\end{figure}

\textbf{Combining with Celeritas.} We combine Celeritas's code for selecting particle energies with our code for calculating directions and writing outputs. This hybrid increases GPU throughput by \textbf{16.1\%} over Celeritas (Table~\ref{tab:hybrid}). It improves GPU throughput over our generated code by \textbf{1.2\%}, measured as the median gain across six paired timing rounds, showing that the two implementations offer complementary improvements and an opportunity for human--AI collaboration.

\begin{table}[!ht]
  \caption{Combining Celeritas components with generated code gives the highest GPU throughput. Times are medians in milliseconds over six paired timing rounds. Throughput gains are medians of per-round time ratios, not ratios of displayed median times. Execution time directly measures transfers, computation, and completion waits in the shared benchmark. Settings are selected separately for each timing metric. Celeritas and the hybrid use a shared direction correction.}
  \centering\small
  \setlength{\tabcolsep}{4pt}
  \renewcommand{\arraystretch}{0.85}
  \begin{tabular}{lrrrr}
    \toprule
    Version & Execution time & GPU time &
      \multicolumn{2}{c}{GPU throughput gain} \\
    & & & vs. Celeritas & vs. CDES \\
    \midrule
    Celeritas & 2.4753 & 0.5093 &  & \\
    CDES-generated & 2.4000 & 0.4433 & \textbf{14.9\%} &  \\
    Hybrid & 2.3905 & \textbf{0.4387} &
      \textbf{16.1\%} & \textbf{1.2\%} \\
    \bottomrule
  \end{tabular}
  \label{tab:hybrid}
\end{table}

\textbf{Ablation.} We compare the evolutionary search with and without certificate feedback, using the same correctness checks (Table~\ref{tab:ablation}). Certificates increase the fraction of proposals that compile and pass these checks from \textbf{55\% to 90\%}. The fraction of proposals that also improve throughput by over 10\% increases from \textbf{50\% to 80\%}. Search without certificates still produces useful candidates at lower total API cost, while certificate feedback improves the passing rate and lowers cost per passing proposal by \textbf{10.7\%}.

\begin{table}[!ht]
  \caption{Certificate feedback produces more proposals that pass correctness checks and improve performance. Costs cover this ablation's Qwen3-Coder-Plus calls.}
  \centering\small
  \setlength{\tabcolsep}{4pt}
  \renewcommand{\arraystretch}{0.85}
  \begin{tabular}{lrr}
    \toprule
    & Without certificates & With certificates \\
    \midrule
    Proposals passing correctness checks & 55\% & \textbf{90\%} \\
    Passing with throughput gain $>10\%$ & 50\% & \textbf{80\%} \\
    Passing with throughput gain $>20\%$ & 30\% & \textbf{35\%} \\
    Best throughput gain on withheld inputs & 66.5\% & \textbf{68.6\%} \\
    Model-call cost (USD) & \$0.0535 & \$0.0782 \\
    \bottomrule
  \end{tabular}
  \label{tab:ablation}
\end{table}

\section{Conclusion and Future Work}

We present a proof-of-concept approach toward reliably and scalably porting large legacy scientific codebases, demonstrated on two Geant4 functions while enforcing a flexible range of correctness requirements. The generated code exhibits optimization patterns that improve on expert-written implementations, and combining it with Celeritas components shows complementary benefits from human–AI collaboration. CDES connects conflict detection, certificate validation, and enforced search control for scientific optimization. Future work will test how the cost and correctness benefits of certificates vary across functions and optimization difficulty, extend evaluation to larger portions of Geant4, and incorporate requirements needed for projects such as Celeritas.

\interlinepenalty=10000
\renewcommand{\bibsection}{\section*{References}}
\bibliographystyle{unsrtnat}
\bibliography{outline_refs}
\end{document}